\documentclass{article}

 \PassOptionsToPackage{numbers, compress}{natbib}

\usepackage[preprint]{neurips_2020}
\usepackage{times}
\usepackage{epsfig}
\usepackage{graphicx}
\usepackage{amsmath}
\usepackage{amssymb}
\usepackage{boldline,multirow} %
\usepackage{subcaption}

\usepackage{amsmath,amsfonts,bm}

\def\eqref#1{equation~\ref{#1}}

\def\1{\bm{1}}

\def\vb{{\bm{b}}}

\def\vt{{\bm{t}}}

\def\vx{{\bm{x}}}

\def\vz{{\bm{z}}}

\def\mA{{\bm{A}}}

\def\mD{{\bm{D}}}

\def\mI{{\bm{I}}}

\def\mK{{\bm{K}}}

\def\mT{{\bm{T}}}

\def\mW{{\bm{W}}}
\def\mX{{\bm{X}}}
\def\mY{{\bm{Y}}}
\def\mZ{{\bm{Z}}}

\DeclareMathAlphabet{\mathsfit}{\encodingdefault}{\sfdefault}{m}{sl}
\SetMathAlphabet{\mathsfit}{bold}{\encodingdefault}{\sfdefault}{bx}{n}

\def\gG{{\mathcal{G}}}

\def\gN{{\mathcal{N}}}

\def\sN{{\mathbb{N}}}

\def\sZ{{\mathbb{Z}}}

\newcommand{\E}{\mathbb{E}}

\newcommand{\R}{\mathbb{R}}

\usepackage[pagebackref=true,breaklinks=true,letterpaper=true,colorlinks,bookmarks=false]{hyperref}

\begin{document}

\title{Infinitely Wide Graph Convolutional Networks: Semi-supervised Learning via Gaussian Processes}

\author{Jilin Hu$^1$, Jianbing Shen$^1$, Bin Yang$^2$, Ling Shao$^1$ \\
$^1$Inception Institute of Artificial Intelligence, Abu Dhabi, UAE\\
$^2$Department of Computer Science, Aalborg University, Denmark\\
\texttt{\{hujilin1229, shenjianbingcg\}@gmail.com}, \\  \texttt{byang@cs.aau.dk}, \texttt{ling.shao@ieee.org} \\}

\maketitle

\begin{abstract}
Graph convolutional neural networks~(GCNs) have recently demonstrated promising results on graph-based semi-supervised classification, but little work has been done to explore their theoretical properties. 
Recently, several deep neural networks, e.g., fully connected and convolutional neural networks, with infinite hidden units have been proved to be equivalent to Gaussian processes~(GPs). 
To exploit both the powerful representational capacity of GCNs and the great expressive power of GPs, we investigate similar properties of infinitely wide GCNs. More specifically, we propose a GP regression model via GCNs~(GPGC) for graph-based semi-supervised learning. In the process, we formulate the kernel matrix computation of GPGC in an iterative analytical form. 
Finally, we derive a conditional distribution for the labels of unobserved nodes based on the graph structure, labels for the observed nodes, and the feature matrix of all the nodes. 
We conduct extensive experiments to evaluate the semi-supervised classification performance of GPGC and demonstrate that it outperforms other state-of-the-art methods by a clear margin on all the datasets while being efficient. 
\end{abstract}

\section{Introduction}
The recent increase in access to large amounts of labeled data, through the development of datasets such as MNIST~\cite{deng2012mnist}, SVHN~\cite{netzer2011reading}, CIFAR~\cite{krizhevsky2010convolutional}, and ImageNet~\cite{russakovsky2015imagenet}, has enabled significant progress in various fields of machine learning, and computer vision in particular.
However, collecting such large-scale labeled datasets remains expensive and time consuming. 
Thus, it is of high interest to develop methods that can automatically annotate extensive amounts of data, when only provided with a small subset of labeled samples. %
To this end, semi-supervised learning~(SSL) has begun to attract significant attention in the machine learning ~\cite{chapelle2009semi,kamnitsas2018semi} and computer vision~\cite{jiang2019semi,luo2018smooth,Iscen_2019_CVPR,Lee_2019_CVPR,Honari_2018_CVPR} communities. 
Current SSL methods can be classified into several categories, including expectation-maximization~(EM) with generative mixture models, self-training methods, co-training methods, transductive support vector machines, and graph-based methods~\cite{pise2008survey}. 

Among these, graph-based methods represent one of the most important categories for following two main reasons. First, data that can be described by a graph structure is ubiquitous in the real world. For example, scene graphs representing relationships among interconnected objects in images, social networks indicating friendships between different users, road networks demonstrating traffic conditions, and so on. 
Second, the embedding graph structure between labeled and unlabeled data provides additional context information to further enhance model performance. In this representation, labeled and unlabeled samples are denoted as nodes in a graph, and the edges between them usually represent their similarities. 
The underlying assumption is that there is a label smoothness over the graph, i.e., the labels of neighbor nodes are similar. 

Many successful graph-based SSL models have been proposed in recent years~\cite{zhu2003semi,belkin2006manifold,weston2012deep,yang2016revisiting,ng2018bayesian,kipf2016semi,gat}. These include graph regularization methods, deep learning based methods, and Gaussian process~(GP) based methods. Early regularization frameworks can be expressed as learning a function $f$ with two terms in the objective function: a loss function for the labeled data, and a graph regularization term. 
As an extension to these early models, deep learning methods have significantly boosted the performance of SSL. In fact, the top-three state-of-the-art methods for SSL across several public graph datasets are deep learning based models, i.e., graph convolutional neural networks~(GCNs)~\cite{kipf2016semi}, mixture model network~(MoNet)~\cite{monti2017geometric}, and graph attention network~(GAT)~\cite{gat}. 

However, deep learning models are lack of theoretical principle~\cite{novak2019bayesian, cnn_as_gp}, which is 
extremely difficult due to their large number of parameters~\cite{cnn_as_gp,wilson2016stochastic}. To fix this problem, the Bayesian neural network was proposed as a single-layer fully connected~(FC) network with infinitely wide hidden units that can be converged to a GP~\cite{neal1996priors}. More efforts have been made to investigate the GP behavior of FC networks with more than one hidden layer~\cite{matthews2018gaussian,lee2018deep}. Further, these ideas have also been transfered to convolutional neural networks, providing their equivalence as GPs~\cite{novak2019bayesian, cnn_as_gp}. Most recently, the relationships between spatial graph convolutions and Gaussian processes were investigated~\cite{walker19a}. 
To the best of our knowledge, no prior work has looked into the relationship between spectral-based GCNs and GPs yet. 

In this paper, we show that GCNs with an infinite number of convolutional filters are equivalent to GPs. Then, we evaluate this equivalence on the task of graph-based semi-supervised classification. 
Our main contributions can be summarized as follows. 
\begin{itemize}
   \item We investigate the GP prior of graph convolutional networks with infinitely wide layers, deriving an equivalence of GCNs as GPs.
  \item We formulate an iterative analytical form of kernel matrix computation for the GP prior by restricting the activation function to be ReLU.
  \item We conduct extensive experiments for graph-based semi-supervised classification on several public datasets, where our proposed methods achieve the state-of-the-art results, outperforming the existing graph-based semi-supervised learning algorithms.
\end{itemize}

\section{Related Work}
\label{related_work}
Graph-based semi-supervised learning has been actively studied in recent years. The most common approaches can be classified into four categories: graph Laplacian regularization based methods~\cite{zhu2003semi,belkin2006manifold,weston2012deep}, graph embedding based methods~\cite{perozzi2014deepwalk,yang2016revisiting,grover2016node2vec,tang2015line}, GP based methods~\cite{lee2018deep,silva2008hidden,chu2007relational}, and deep learning based methods~\cite{kipf2016semi,monti2017geometric,atwood2016diffusion}. 

Graph Laplacian regularization based methods are built on the assumption that labels vary smoothly along the node connections in a graph. As such, a Laplacian regularization is added to the objective function, which needs to be minimized. The object functions of these methods are typically constructed from either a Gaussian field~\cite{zhu2003semi} or certain kernel functions~\cite{belkin2006manifold}. 
Recently, Weston et al.~\cite{weston2012deep} proposed a deep learning model that includes the embedding instances of output labels, hidden layers, and an auxiliary layer as regularizers. However, none of these methods are competitive with currently state-of-the-art methods. 

Graph embedding based methods are mostly derived from the  Skipgram~\cite{frome2013devise} model, and are formulated to predict the context of a node by taking the embedding features as input.  
Deepwalk~\cite{perozzi2014deepwalk} aims to learn node embeddings by exploiting the graph topology information with random walks. Since this process is a form of unsupervised learning, the embedding features can be further utilized by other classification algorithms. LINE~\cite{tang2015line} is an extension of Deepwalk that considers multiple context spaces. 
However, none of these methods leverage label information, making them unsuitable for the task of graph-based semi-supervised classification. In contrast, Planetoid~\cite{yang2016revisiting} proposed a semi-supervised learning framework based on graph embedding, which obtained competitive results. 

Recently, more and more deep learning models have been proposed to better address the task of graph semi-supervised classification. Atwood and Towsley~\cite{atwood2016diffusion} proposed a diffusion convolutional neural network that considers the diffusion process on a graph, and better captures the graph structured data than fully-connected neural networks. Graph convolutional neural networks~(GCNs)\cite{bruna2013spectral,defferrard2016convolutional,kipf2016semi}, which are derived from the graph spectral theory, have also received significant attention, reaching state-of-the-art performance in semi-supervised learning on graph structured data. More recently, Monti et al. proposed a generic spatial domain framework for deep learning on non-Euclidean domains such as graphs and manifolds\cite{monti2017geometric}, which has also achieved competitive results. However, a common problem within deep learning methods is that they require a relatively large labeled validation dataset in order to allow early stopping during training to prevent over-fitting~\cite{ng2018bayesian}. 

Therefore, a graph Gaussian processes~(GGP) method was proposed, which is a data-efficient Gaussian process model and does not need large validation data~\cite{ng2018bayesian}. This GP-based method provided a Bayesian approach to address the semi-supervised learning problem and obtained competitive results compared to GCNs, giving it a big advantage over neural network based methods. However, it is limited in considering only 1-hop neighborhood nodes whose weights are equal, thus ignoring their differences.

\section{Preliminaries}
\label{bg}
In this section, we briefly review the theoretical motivations behind fast approximate graph convolutions~(Section~\ref{sec:fagc}) and infinitely wide neural networks as GPs~(Section~\ref{sec:nngp}), of which we will take advantage to formalize our model in Section~\ref{gcn_gp}. 

\subsection{Notation}
\label{sec:notation}
Let $\mX^{(0)} \in \R^{N\times m^{(0)}}$ and $\mA \in \R^{N\times N}$ denote the graph-structured input feature matrix and the adjacency matrix of underlying graph $\gG$, respectively, where $N$ is the total number of nodes and $m^{(0)}$ is the input feature dimension.
Suppose we have a graph convolutional neural network with a total of $L$ layers. The layer-wise formulation can be represented as the following two steps: 1) $\mZ^{(l)} = g_{\theta} \star \mX^{(l-1)}$, where $g_{\theta}$ denotes the graph convolution filter parametrized by weight matrix $\mW^{(l)}$, and $\star$ denotes the graph convolution operation; and 2) $\mX^{(l)} = \Phi(\mZ^{(l)})$, where $\mZ^{(l)} \in \R^{N\times m^{(l)}}$ and $\mX^{(l)} \in \R^{N\times m^{(l)}}$ denote the pre- and post-activated output at the $l$-th layer, $\Phi(\cdot)$ is the activation function, and $m^{(l)}$ is the corresponding feature dimension of the $l$-th layer. %

\subsection{Fast Approximate Graph Convolution}
\label{sec:fagc}
The fast approximate graph convolution is derived from the spectral based graph convolution by applying several approximations~\cite{kipf2016semi}. The resulting graph convolution operation at the $l$-th layer can be formulated as follows,
\begin{equation}
\mX^{(l)} = \Phi(\mZ^{(l)})
\label{eq:gcn_nl}
\end{equation}
\begin{equation}
\mZ^{(l)} =\widetilde{\mD}^{-1/2}\widetilde{\mA}\widetilde{\mD}^{-1/2} \mX^{(l-1)} \mW^{(l)},
\label{eq:fgcn}
\end{equation}
\noindent
where $\widetilde{\mA} = \mA + \mI$ is the adjacency matrix with self-loops, $\widetilde{\mD}$ is the degree matrix, where $\widetilde{\mD}_{i,i} = \sum_j\widetilde{\mA}_{i,j}$, and $\widetilde{\mD}_{i,j} = 0$ for $i \neq j$. Let $\hat{\mA} = \widetilde{\mD}^{-1/2}\widetilde{\mA}\widetilde{\mD}^{-1/2}$. Equation~\ref{eq:fgcn} can be simplified as: 
\begin{equation}
\mZ^{(l)} =\hat{\mA} \mX^{(l-1)} \mW^{(l)}.
\label{eq:fgcn_sim}
\end{equation}
Since $\mA$ is fixed once the graph $\gG$ is specified, $\hat{\mA}$ is a constant matrix.

One simplified example of a two-layer GCN for semi-supervised classification was proposed in~\cite{kipf2016semi} as,
\begin{equation}
\mZ^{(2)} =softmax(\hat{\mA}\Phi(\hat{\mA} \mX \mW^{(0)})\mW^{(1)}),
\label{eq:eg_gcn}
\end{equation}
where $\mZ^{(2)}\in\R^{N\times C}$ is the output, with $C$ being the total number of classes, and $\Phi(\cdot)$ is selected as the ReLU function.  

\subsection{Infinitely Wide Neural Networks as GPs}
\label{sec:nngp}
Given the input $\vx^{(0)} \in \R^{m^{(0)}}$ and output $\vz^{(1)} \in \R^{m^{(1)}}$, a single layer fully-connected neural network can be formulated as, 
\begin{equation}
\vx_{j}^{(1)} = \Phi(\sum_{k=1}^{m^{(0)}}\mW_{j,k}^{(0)}\vx_{k}^{(0)}+\vb_{j}^{(0)}), 1\leq j \leq L_1
\label{eq:nn_1}
\end{equation}
\begin{equation}
\vz_{i}^{(1)} = \sum_{j=1}^{L_1}\mW_{i,j}^{(1)}\vx_{j}^{(1)} +\vb_{i}^{(1)}, 1\leq i\leq m^{(1)},
\label{eq:nn_2}
\end{equation}
\noindent
where $\vx_{j}^{(1)}$ is the $j$-th component of the hidden layer, $\vz_{i}^{(1)}$ is the $i$-th component of the output, and $\mW_{i,j}^{(1)} \sim \gN(0, \sigma_{\mW^{(1)}}^{2})$ and $\vb_{i}^{(1)} \sim \gN(0, \sigma_{\vb^{(1)}}^{2})$ are the weight and bias parameters, which are independent. 
Moreover, since $\forall \vx^{(1)}_{j} \in \vx^{(1)}$ is bounded, the variance for $\vx^{(1)}_{j}$ is finite. 
According to the Central Limit Theory, when the width of the hidden layer goes to infinity, i.e. $L_1 \rightarrow +\infty $, $\vz_{i}^{(1)}$ follows a Gaussian distribution~\cite{neal1996priors}. 

When we have multiple inputs $\mT_f = \{\vt_1, \cdots, \vt_f\}$, where $\vt_j \in \R^{m^{(0)}}, 1\leq j \leq f$, the corresponding outputs of the $i$-th component $\sZ_i^{(1)} = \{\vz_i^{(1)}(\vt_1), \cdots, \vz_i^{(1)}(\vt_f)\}$ follow a joint multivariate Gaussian distribution~\cite{lee2018deep}, which can be represented as: 
\begin{equation}
P(\sZ_i^{(1)}|\mT_f) \sim GP(\mu^{(1)}, \Theta^{(1)}), \notag
\end{equation}
where $\mu^{(1)}\in \R^{f}$ is the mean vector and $\Theta^{(1)} \in \R^{f\times f}$ is the covariance matrix~\cite{lee2018deep}. Recalling that $\vb_{j}^{(1)}$ and $\mW_{j,k}^{(1)}$ have zero means and are i.i.d, we draw the following conclusions: 
1) $\mu^{(1)}=0$, and 2) 
\begin{align}
\begin{split}
\label{eq:cov_1}
&\Theta^{(1)}_{m,n}(\vt_m, \vt_n)\\
=&\E[ \vz_i^{(1)}(\vt_m) \vz_i^{1}(\vt_n) ] \\
=&\sigma_{\vb^{1}}^{2} + \sum_{j}{\sigma_{\mW^{(1)}}^{2}\E [x_j^{(1)}(\vt_m)x_j^{(1)}(\vt_n)]} \\
=&\sigma_{\vb^{(1)}}^{2} + L_1{\sigma_{\mW^{(1)}}^{2}\E [x_j^{(1)}(\vt_m)x_j^{(1)}(\vt_n)]},
\end{split}
\end{align}
\noindent
where $\mu^{(1)}$ and $\Theta^{(1)}$ are the same for all output dimensions, since the outputs at different dimensions, i.e. $\sZ_{i}^{(1)}$ and $\sZ_{j}^{(1)}$, $i \neq j$, are independent~\cite{lee2018deep,neal1996priors}. 

Similar conclusions can be extended to deep neural networks with $L$ layers as GPs~\cite{lee2018deep}, whose $i$-th component of the output can be represented as:
\begin{equation}
P(\sZ_i^{(L)}|\mT_f) \sim GP(0, \Theta^{(L)}).
\label{eq:gp_output}
\end{equation}
$\Theta^{(L)}$ can be computed iteratively by following the shorthand quantity relationship as indicated in~\cite{lee2018deep},
\begin{align}
\begin{split}
\label{eq:cov_induction}
&\Theta^{(l)}_{m,n}(\vt_m, \vt_n) \\
=& \sigma_{\vb^{(l)}}^{2} + \delta_\mW T_{\Phi}[\Theta^{(l-1)}_{n,n}(\vt_n, \vt_n), \Theta^{(l-1)}_{m,n}(\vt_m, \vt_n),\\
&\Theta^{(l-1)}_{m,m}(\vt_m, \vt_m)],
\end{split}
\end{align}
\noindent
where $T_{\Phi}(\cdot)$ is a deterministic function that is only related to the activation function $\Phi(\cdot)$. 
It is worth noting that $L_1\sigma_{\mW^{(1)}}^{2}$ in Equation~\ref{eq:cov_1} approaches infinity when $L_1$ goes to infinity and $\sigma_{\mW^{(1)}}^{2}$ is a fixed value. To avoid this, 
we assume that $\delta_\mW = L_{l}\sigma_{\mW^{(l)}}^{2}$ is a \textbf{fixed constant value} across different layers $l$~\cite{neal1996priors,matthews2018gaussian}. The initial covariance matrix is given by $\Theta^{(0)}_{m,n}(\vt_m, \vt_n) = \sigma_{\vb^{(0)}}^{2}  + \delta_\mW K_{\theta}(\vt_m,\vt_n)$, where $K_{\theta}(\vt_m,\vt_n)$ is the initial kernel function, which can be selected from a set of existing, well-defined kernels~\cite{duvenaud2014automatic}.

When the activation is set to \emph{ReLU}, an analytical form of the covariance matrix can be derived via the arc-cosine kernel~\cite{cho2009kernel,lee2018deep}, which is described as follows,
\begin{align}
\begin{split}
\label{eq:acos_kernel}
 &\Theta^{(l)}_{m,n}(\vt_m, \vt_n) \\
= & \sigma_{\vb^{(l)}}^{2} + \frac{\delta_\mW}{\pi}\sqrt{\Theta^{(l-1)}_{n,n}(\vt_n, \vt_n) \Theta^{(l-1)}_{m,m}(\vt_m, \vt_m) }(\text{sin}\alpha_{m, n}^{(l-1)} + \\
 & (\pi - \alpha_{m, n}^{(l-1)})\text{cos}\alpha_{m,n}^{(l-1)})
\end{split}
\end{align}
\begin{equation}
\label{eq:acos_theta}
\alpha_{m, n}^{(l)} = {\text{cos}}^{-1}(\frac{\Theta^{(l)}_{m,n}}{\sqrt{\Theta^{(l)}_{m,m}\Theta^{(l)}_{n,n}}}).
\end{equation}

\section{Gaussian Processes and Graph Convolution Networks}
\label{gcn_gp}
In this section, we describe the Gaussian processes for infinitely wide layers in GCNs. To clarify this process, we first decompose the GCN operations~(Equation~\ref{eq:gcn_nl} and~\ref{eq:fgcn_sim}) on the $l$-th layer into the following three steps: 
\begin{equation}
\label{eq:gcn_1}
\mX_{i,j}^{(l-1)} =  \Phi(\mZ_{i,j}^{(l-1)})
\end{equation}
\begin{equation}
\hat{\mX}_{i,j}^{(l-1)} = \sum_{k=1}^{N}\hat{\mA}_{i,k} \mX_{k,j}^{(l-1)}
\label{eq:adj_transform}
\end{equation}
\begin{equation}
\label{eq:gcn_3}
\mZ_{i,j}^{(l)} =  \sum_{k=1}^{L_{l}}\hat{\mX}_{i,k}^{(l-1)} \mW_{k,j}^{(l)},
\end{equation}
where $\mX_{i,j}^{(l-1)}$ and $\mZ_{i,j}^{(l)}$ are the input and output of node $i$, $j$-th component on the $l$-th layer, respectively, and $\mW^{(l)}_{k,j} \sim N(0, \sigma_{\mW^{(l)}}^{2})$ are the weight parameters. 

We note that Equations~\ref{eq:gcn_1}, \ref{eq:adj_transform} and \ref{eq:gcn_3} are very similar to Equations~\ref{eq:nn_1} and~\ref{eq:nn_2}, except for the additional operation in Equation~\ref{eq:adj_transform}. Moreover, we also observe that this operation still maintains the independent relationship between $\hat{\mX}^{(l-1)}_{n, i}$ and $\hat{\mX}^{(l-1)}_{n, j}$, $\forall i \neq j$. Therefore, the conclusion in Equation~\ref{eq:gp_output} can be derived as,
\begin{equation}
\label{eq:ggp_prior}
P(\mZ_i^{(L)}|\mX, \mA) \sim GP(0, \Gamma^{(L)}),
\end{equation}
where $\Gamma^{(L)}$ is the co-variance matrix of outputs. The co-variance matrix on the $l$-th layer can be reformulated from Equation~\ref{eq:cov_induction} as,
\begin{equation}
\label{eq:gcn_f}
\Gamma^{(l)}_{m,n}(\mX_m, \mX_n)=\delta_\mW\sum_{i}^{N}\sum_{j}^{N}{[(\hat{\mA}_{m\cdot}^{T} \times \hat{\mA}_{n\cdot})\odot \Delta^{(l)}]_{i,j}}, 
\end{equation}
\noindent
where 
$\Delta^{(l)}_{m,n}(\mX_m, \mX_n)=T_{\Phi}[\Gamma^{(l-1)}_{n,n}(\mX_n, \mX_n)$, $\Gamma^{(l-1)}_{m,n}(\mX_m, \mX_n)$, $ \Gamma^{(l-1)}_{m,m}(\mX_m, \mX_m)]$ denotes the covariance matrix over the non-linear activation function $\Phi(\cdot)$, $\hat{\mA}_{m\cdot}, \hat{\mA}_{n\cdot} \in \R^{1\times N}$ denote the $m$-th and $n$-th row vectors of $\hat{\mA}$, respectively, and $\odot$ denotes an element-wise multiplication.

\subsection{Semi-supervised Classification with Gaussian Process}
\begin{figure*}[tbp]
\begin{center}
	\includegraphics[width=0.9\linewidth]{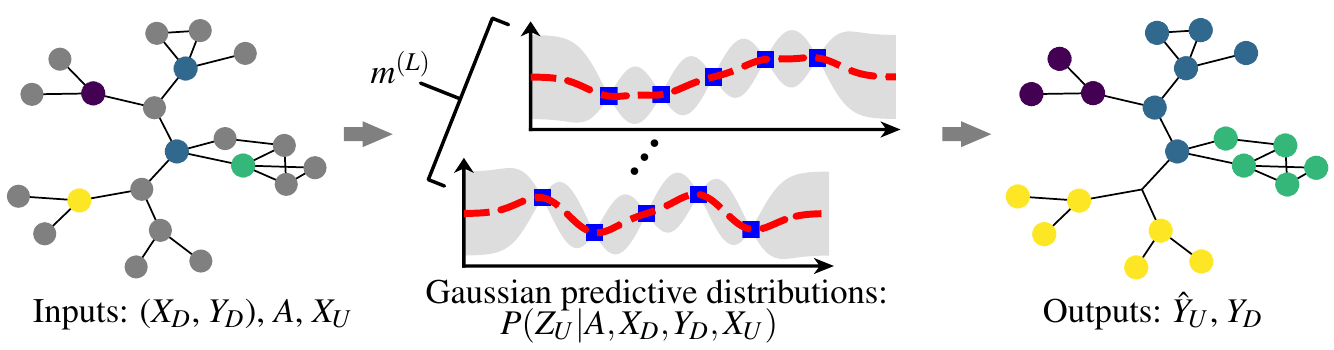}
\caption{The Framework of GPGCs. \emph{Left}: The grey nodes denote the unlabeled data with feature matrix $\mX_U$, while the colored nodes indicate the labeled data, $\mX_D$, and the corresponding colors denotes the labels, $\mY_D$, which constitute the observed data pairs $(\mX_D, \mY_D)$. The embedding graph structure is denoted by the adjacency matrix $\mA$. \emph{Middle}: Description of the resulting GP distribution conditioned on the observations. \emph{Right: }The classification results $\hat{\mY}_U$ on unobserved nodes, and $\mY_D$ for labeled nodes. }
\label{fig:framework}
\end{center}
\end{figure*}

We introduce the process of proceeding semi-supervised classification with Gaussian process regression. Given a graph $\gG$ with an adjacency matrix $\mA$ and feature matrix $\mX\in \R^{N\times m^{(0)}}$ for all nodes, a set of labeled nodes $\sN_D=\{n_1, \cdots, n_D\}$ with feature matrix $\mX_D$ and label matrix $\mY_D$,  where $D\ll N$. Thus, we have a large set of unobserved nodes in $\sN_U=\{n_{D+1}, \cdots, n_N\}$, whose feature matrix is $\mX_U\in \R^{(N-D)\times m^{(0)}}$. Let $\mZ_D$ and $\mZ_U$ be the outputs of an $L$-layer GCNs for the inputs $\mX_D$ and $\mX_U$, respectively. We aim to derive the best node label estimation $\hat{\mY}_U$ for $\sN_U$. Figure~\ref{fig:framework} depicts the framework of GPGC.

According to Equation~\ref{eq:ggp_prior}, the Gaussian process prior for $\mZ_D$ and $\mZ_U$ can be represented as a joint Gaussian distribution:
\begin{align}
\begin{split}
 P(\mZ_D, \mZ_U|\mX_D,\mX_U, \mA) 
\sim GP\left(0,
\left[
\begin{array}{cc}
 \Gamma_{D, D}^{L}  &   (\Gamma_{U, D}^{L})^{T}  \\
  \Gamma_{U, D}^{L}  &    \Gamma_{U, U}^{L}
\end{array}
\right]
\right), %
\label{eq:joint}
\end{split}
\end{align}
where $\Gamma_{D, D}^{L} \in \R^{D\times D}$ is the covariance matrix over labeled data, $\Gamma_{U, D}^{L} \in \R^{(N-D)\times D}$ is between unlabeled and labeled data, and $\Gamma_{U, U}^{L} \in \R^{(N-D)\times (N-D)}$ is over unlabeled data. 

However, we only have access to the label matrix $\mY_D$, which can be represented as $\mY_D = \mZ_D + \tau$, where $\tau$ is the observation noise. We further assume that the noise $\tau$ is i.i.d with zero mean and variance $\sigma_{\tau}^{2}$. Thus, the covariance relationship between $\mY_D$ and $\mZ_D$ can be formulated as $K(\mY_D, \mY_D) = \Gamma_{D, D}^{L} + \sigma_{\tau}^{2}\mI_{D}$, where $\mI_D\in \R^{D\times D}$ is an identity matrix. Therefore, the joint distribution of $\mY_D$ and $\mZ_U$ can be derived from Equation~\ref{eq:joint} as,
\begin{align}
\begin{split}
P(\mY_D, \mZ_U|\mX_D,\mX_U, \mA) 
 \sim GP\left(0,
\left[
\begin{array}{cc}
 \Gamma_{D, D}^{L} + \sigma_{\tau}^{2}\mI_{D} &   (\Gamma_{U, D}^{L})^{T}  \\
  \Gamma_{U, D}^{L}  &    \Gamma_{U, U}^{L}
\end{array}
\right]
\right). %
\label{eq:mean_gp}
\end{split}
\end{align}

As is standard in GPs, we have the conditional distribution~\cite{williams2006gaussian}, 
$P(\mZ_U|\mA, \mX_D, \mX_U, \mY_D) \sim \gN(\bar{\mZ}_U, \Lambda)$, where the mean matrix $\bar{\mZ}_U\in \R^{(N-D) \times m^{(L)}}$ and covariance matrix $\Lambda \in \R^{(N-D) \times (N-D)}$ are formulated as follows,
\begin{equation}
\bar{\mZ}_U = \Gamma_{U, D}^{L} (\Gamma_{D, D}^{L} + \sigma_{\tau}^{2}\mI_D)^{-1}\mY_D \notag
\end{equation}
\begin{equation}
\Lambda = \Gamma_{U, U}^{L} - \Gamma_{U, D}^{L} (\Gamma_{D, D}^{L} + \sigma_{\tau}^{2}\mI_D)^{-1}(\Gamma_{U, D}^{L})^{T}.\notag
\end{equation}

Finally, the labels $\hat{\mY}_U\in \R^{(N-D)}$ can be obtained via 
\begin{equation}
[\hat{\mY}_U]_{i} = \text{argmax}([\bar{\mZ}_U]_{i\cdot}), 1\leq i \leq N-D,\notag
\end{equation}
where $[\bar{\mZ}_U]_{i\cdot}$ is the $i$-th row of $\bar{\mZ}_U$, i.e. the mean of the estimations for node $i$, and $\text{argmax}(\cdot)$ returns the index of the largest value in $[\bar{\mZ}_U]_{i\cdot}$. Thus $[\hat{\mY}_U]_{i}$ denotes the estimation label for node $i$. 

\subsection{GPs with Two-Layer GCNs}
\label{sec:eg_gcgp}
In this paper, we consider a GP with two-layer GCNs formulated in Equation~\ref{eq:eg_gcn}. To provide a more detailed description, we decompose it into the following five steps: $1) \hat{\mX}^{(0)} = \hat{\mA} \mX; 2) \mZ^{(1)} =  \hat{\mX}^{(0)} \mW^{(1)}; 3) \mX^{(1)} =  \Phi(\mZ^{(1)}); 4) \hat{\mX}^{(1)} = \hat{\mA} \mX^{(1)}; 5) \mZ^{(2)} =  \hat{\mX}^{(1)} \mW^{(2)}$.
We note that it is quite time consuming to compute the covariance matrix in Step 1, which will be discussed in Section~\ref{sect:runtime}. However, we also find that this operation does not change the column dependency of $\mX$, so it can be treated as a pre-processing operation on the input to ease the computation. 
Then, Steps 2 and~3 consist of a single-layer neural network with a ReLU activation function, whose resulting kernel can be derived from Equations~\ref{eq:acos_kernel} and~\ref{eq:acos_theta} as,
\begin{equation}
\label{eq:acos_theta1}
\alpha_{m, n}^{(0)} = {\text{cos}}^{-1}(\frac{\mK_{m,n}(\hat{\mX}^{(0)}, \hat{\mX}^{(0)})}{\sqrt{\mK_{m,m}(\hat{\mX}^{(0)}, \hat{\mX}^{(0)})\mK_{n,n}(\hat{\mX}^{(0)}, \hat{\mX}^{(0)})}})
\end{equation}
\begin{align}
\begin{split}
\label{eq:acos_kernel1}
&\mK_{m,n}(\mX^{(1)}, \mX^{(1)}) \\
 = &\frac{\delta_{\mW}}{\pi}\sqrt{\mK_{m,m}(\hat{\mX}^{(0)}, \hat{\mX}^{(0)}) \mK_{n,n}(\hat{\mX}^{(0)}, \hat{\mX}^{(0)}) }(\text{sin}\alpha_{m, n}^{(0)} + \\& (\pi - \alpha_{m, n}^{(0)})\text{cos}\alpha_{m,n}^{(0)}).
\end{split}
\end{align}

Finally, the kernel matrix for $\mZ^{(2)}$ can be obtained from Equation~\ref{eq:gcn_f} as,
\begin{align}
\begin{split}
\label{eq:gcn_f1}
 \Gamma^{(2)}_{m,n}(\mX_m, \mX_n) 
=\delta_{\mW}\sum_{i}\sum_{j}{[(\hat{\mA}_{m\cdot}^{T} \times \hat{\mA}_{n\cdot})\odot \mK(\mX^{(1)}, \mX^{(1)}) ]_{i,j}}.
\end{split}
\end{align}

\section{Experimental Results}
In this section, we first introduce our experimental setup, including datasets and baseline methods. 
Then, we evaluate our model for semi-supervised classification on citation and image datasets, and demonstrate that it outperforms the state-of-the-art methods under the same experimental settings. Finally, a detailed discussion of the results is presented. 

\subsection{Experimental Setup}
\noindent
\textbf{Citation Datasets.}
For ease of comparison, we follow exactly the same experimental settings and data splits as indicated in~\cite{kipf2016semi} and~\cite{ng2018bayesian}. We conduct our experiments on three citation datasets~(Cora, Citeseer, and Pubmed~\cite{sen2008collective}), which are reported in Table~\ref{tb:ds}, where \emph{\#Nodes} denotes the total number of documents in the dataset, \emph{\#Edges} denotes the total number of connections between documents, \emph{\#Classes} denotes the number of classes that these documents can be classified into~(one document can only have one class), \emph{\#Features} indicates the length of the Bag-of-Words~(BOW) feature vector for each document, \emph{\#Train} denotes the number of labeled nodes per class that are selected as training data, and \emph{\#Validation} and \emph{\#Test} denote the total number of validation and testing nodes, respectively. From the column \emph{Label Rate}, we can observe that the training label rate can be as low as 0.3\% in Pubmed, which makes the task challenge. 

\begin{table*}[tp]
\caption{Dataset Descriptions}
\label{tb:ds}
\begin{center}
\begin{tabular}{ccccccccc}
\clineB{1-9}{2.5}
Dataset & \#Nodes & \#Edges & \#Classes & \#Features & \#Train & \#Validation &\#Test & Label Rate \\ \hline
Citeseer & 3,327 & 4,732 & 6 & 3,703 & 20/class & 500 & 1000 & 0.036\\
Cora & 2,708 & 5,429 & 7 & 1,433 & 20/class & 500 & 1000 & 0.052 \\
Pubmed & 19,717 & 44,338 & 3 & 500 & 20/class & 500 & 1000 & 0.003 \\ \hline
MNIST & 10,000 & 65,403 & 10 & 128 & 100/class & 1000 & 8000 & 0.1 \\
SVHN & 10,000 & 68,844 & 10 & 128 & 100/class & 1000 & 8000 & 0.1 \\
CIFAR10 & 10,000 & 70,391 & 10 & 128 & 100/class & 1000 & 8000 & 0.1 \\ \clineB{1-9}{2.5}
\end{tabular}
\end{center}
\label{default}
\end{table*}%

\begin{figure*}[tp]
\begin{center}
	\begin{subfigure}{0.28\linewidth}
		\includegraphics[ width=\linewidth]{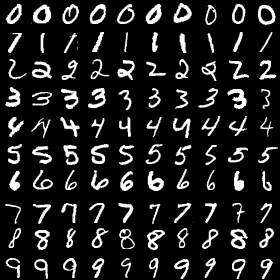}
		\caption{MNIST}
		\label{fig:mnist_origin}
	\end{subfigure}
	\begin{subfigure}{0.28\linewidth}
		\includegraphics[ width=\linewidth]{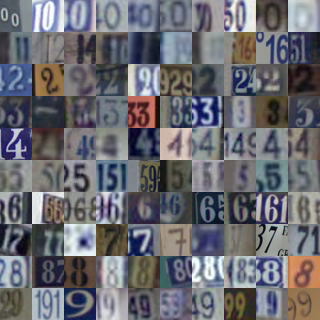}
		\caption{SVHN}
		\label{fig:svhn_origin}
	\end{subfigure}
	\begin{subfigure}{0.28\linewidth}
		\includegraphics[ width=\linewidth]{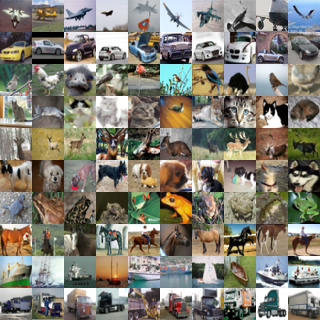}
		\caption{CIFAR10}
		\label{fig:cifar10_origin}
	\end{subfigure}
	
	\begin{subfigure}{0.28\linewidth}
		\includegraphics[trim={2cm 3cm 2cm 3cm},clip, width=\linewidth]{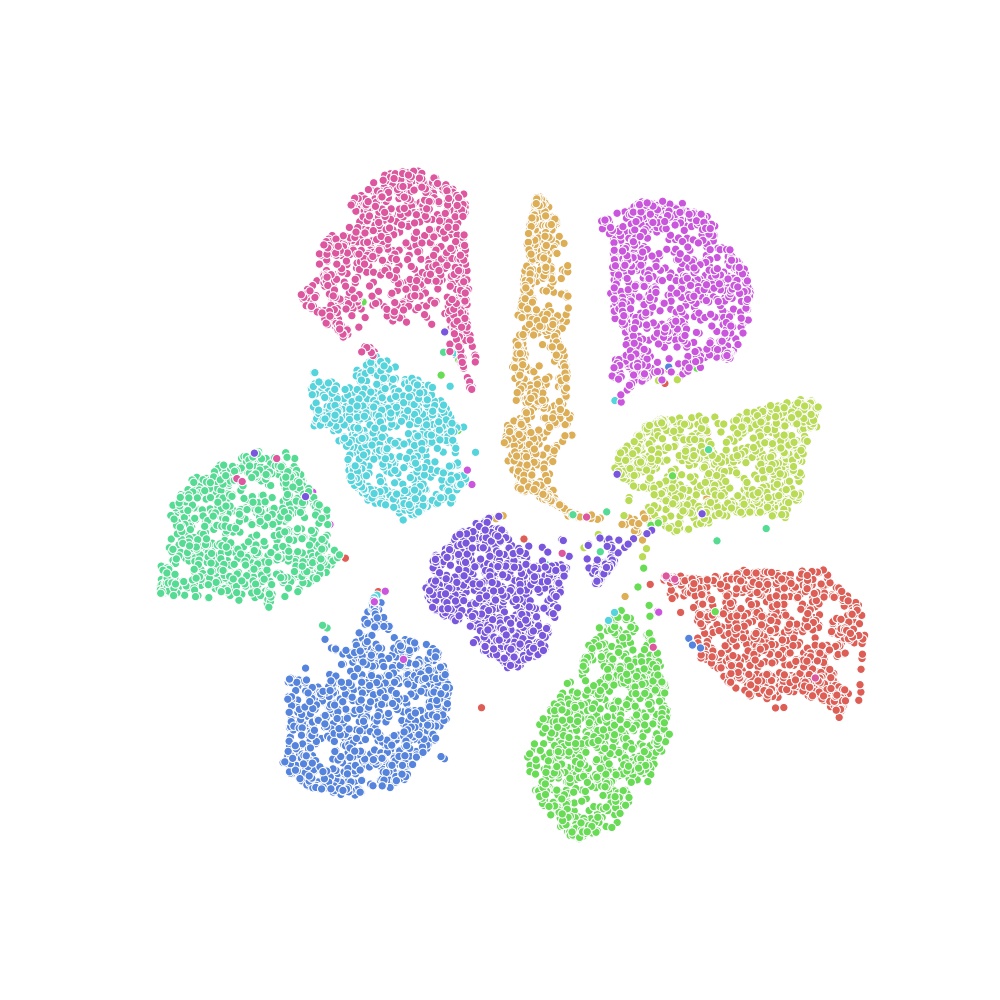}
		\caption{MNIST}
		\label{fig:mnist}
	\end{subfigure}
	\begin{subfigure}{0.28\linewidth}
		\includegraphics[trim={2cm 3cm 2cm 3cm},clip, width=\linewidth]{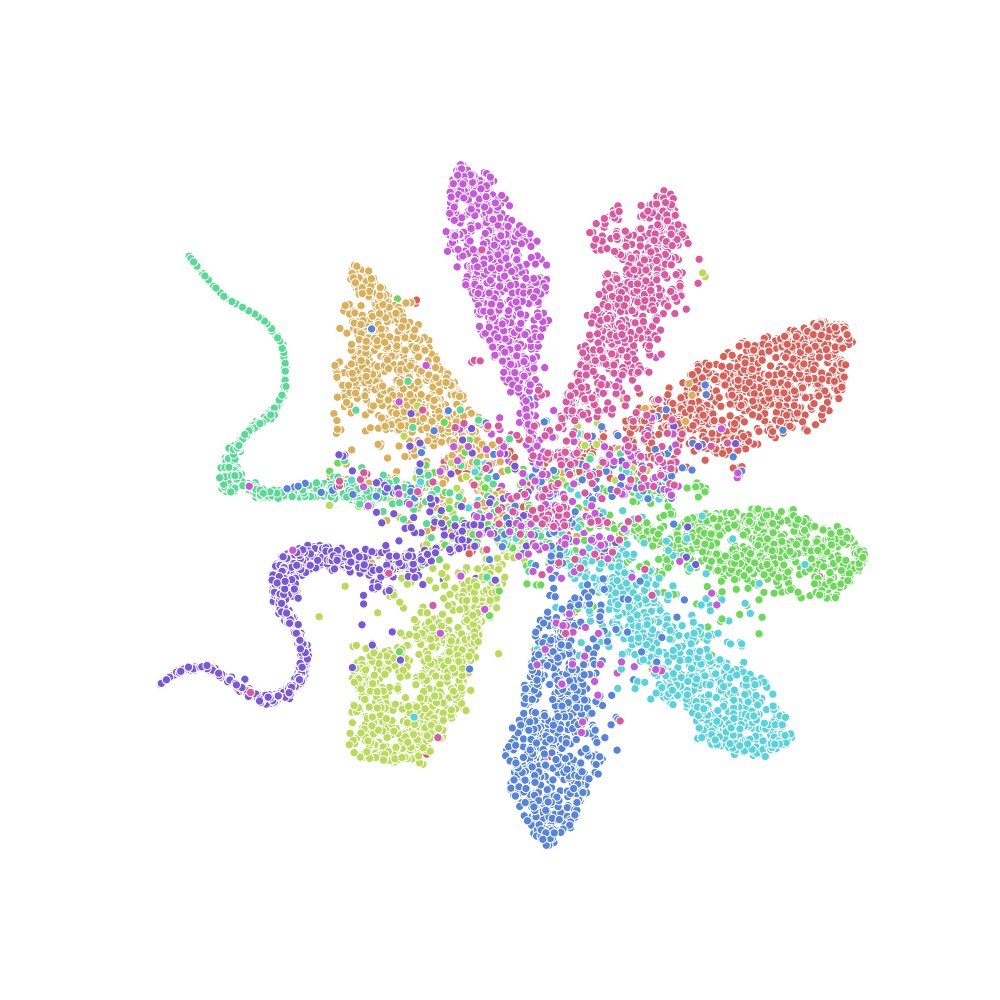}
		\caption{SVHN}
		\label{fig:svhn}
	\end{subfigure}
	\begin{subfigure}{0.28\linewidth}
		\includegraphics[trim={2cm 3cm 2cm 3cm},clip, width=\linewidth]{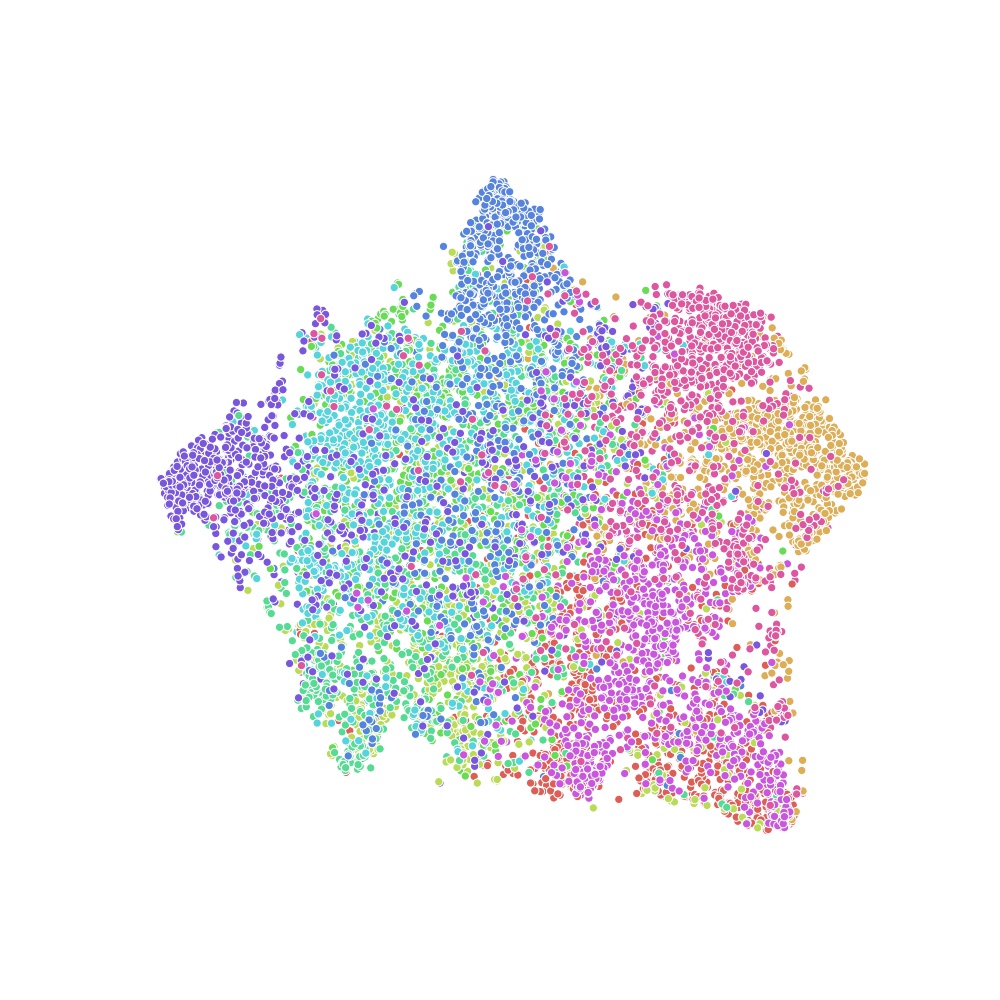}
		\caption{CIFAR10}
		\label{fig:cifar10}
	\end{subfigure}
\caption{Data Representation and t-SNE of MNIST, SVHN, and CIFAR10}
\label{fig:tsne_image}
\end{center}
\end{figure*}

\noindent
\textbf{Image Datasets.} We further evaluate our proposed methods on three widely adopted image datasets: MNIST~\cite{deng2012mnist}, SVHN~\cite{netzer2011reading}, and CIFAR10~\cite{krizhevsky2010convolutional}. 
We randomly sample 1000 images per class for each dataset. Since all three datasets contain 10 classes, we obtain a total of 10,000 images for each. Then, we randomly select 100 images per class for training, 100 for validation, and the remaining 800 for testing. Next, we adopt the model \emph{Conv-Large}, originally proposed in virtual adversarial training~(VAT)~\cite{vat}, to carry out feature extraction, where the final layer with 128 hidden units is output as the finalized feature in our paper. The original data representations and 2-D t-SNE visualizations~\cite{maaten2008visualizing} of features extracted for the three datasets are shown in Figure~\ref{fig:tsne_image}. In Figures~\ref{fig:mnist_origin},~\ref{fig:svhn_origin}, and~\ref{fig:cifar10_origin}, each row contains 10 image samples that have the same label. As can be seen from these data representations, the classification task becomes more challenge from MNIST, to SVHN, to CIFAR10. This is consistent with what we find from the t-SNE representations, shown in Figures~\ref{fig:mnist},~\ref{fig:svhn}, and~\ref{fig:cifar10}.

To construct a graph of the selected images, we treat each image as a node, and compute the pair-wise Euclidean distances between two nodes based on the extracted features. We then select the top-$k$ nearest images as the neighbor nodes, i.e., $k$ edges are constructed between the selected image node and the top-$k$ images.
To ensure that the graph is undirected, we connect nearest neighbor nodes. In this paper, we select $k=10$. The statistics of the image datasets are reported in Table~\ref{tb:ds}.

\noindent
\textbf{Experimental Design.} In Section~\ref{sec:eg_gcgp}, we provide a detailed description for computing the kernel matrix in five steps, and also notice the most time consuming step~(Step 4). Therefore, we propose two models: 1) GPGC-\emph{small} which consists of steps 1, 2, 3, and 5; and 2) GPGC-\emph{big}, which includes all five steps.
Moreover, similar to GGP-X proposed in~\cite{ng2018bayesian}, our proposed method does not need validation data for early stopping, either. Thus, once the optimal hyper-parameters are determined from the validation dataset, we can include validation data into our model training, which can further improve our model's performance, which is denote as~\emph{GPGC-X}. %

Recall the initial kernel function $K_{\theta}(\vt_m,\vt_n)$  mentioned in Section~\ref{sec:nngp}. In our experiments, we consider the following four kernel functions, Squared Exponential~($K_{SE}^{l} = \exp{\left(-\frac{(\vt_m - \vt_n)^{2}}{2l^2}\right)}$), Inner Product~($K_{IP} = \vt_m^{T} \vt_n$), Arccosine~($K_{AC} = 1 - \frac{1}{\pi}\text{cos}^{-1}(\frac{\vt_m^{T} \vt_n}{||\vt_m||\cdot||\vt_n||})$), and Polynomial~($K_{PL} = (\frac{\vt_m^{T} \vt_n}{||\vt_m||\cdot||\vt_n||} + b)^a$), where $b$ and $l$ serve as hyper-parameters in the polynomial and squared exponential kernel, respectively.

Therefore, we have at most three hyper-parameters, $\delta_{\mW}$, $l$ in $K_{SE}^{l}$, and $b$ in $K_{PL}^{b}$, which need to be tuned in all our experiments. 
The best hyper-parameters are found by grid search, and are reported in Table~\ref{tb:hp}.

\begin{table}[tbp]
\caption{Best Hyper-parameters for GPGCs}
\label{tb:hp}
\begin{center}
\begin{tabular}{c|c|ccc}
\clineB{1-5}{2.5}
 & Parameters & Cora & Citeseer & Pubmed \\ \hline
\multirow{2}{*}{GPGC-\emph{small}} & $\delta_{\mW} $ & 0.010 & 0.017 & 0.075 \\
						          & $K_\theta$ & $K_{AC}$ & $K_{IP}$ & $K_{PL}^{0.5}$ \\ \clineB{1-5}{2.5}
\multirow{2}{*}{GPGC-\emph{big}} & $\delta_{\mW} $ & 0.137 &  0.082 &  0.35 \\
						& $K_\theta$  & $K_{AC}$ & $K_{IP}$ & $K_{PL}^{0.5}$  \\ \clineB{1-5}{2.5}
\end{tabular}
\end{center}
\label{default}
\end{table}

\noindent
\textbf{Baselines.} We compare our method against the same baseline methods as in~\cite{ng2018bayesian}; namely, graph Laplacian regularization based methods~(label propagation~(LP)~\cite{zhu2003semi}, semi-supervised embedding~(SemiEmb)~\cite{weston2012deep} and manifold regularization~(ManiReg)~\cite{belkin2006manifold}), embedding based models~(DeepWalk~\cite{perozzi2014deepwalk} and Planetoid~\cite{yang2016revisiting}), deep learning based methods~(DCNN~\cite{yang2016revisiting}, GCNs~\cite{kipf2016semi}, MoNet~\cite{monti2017geometric}, and \textbf{GAT}~\cite{gat}), and the Gaussian process based method~(GGP~\cite{ng2018bayesian}), where \textbf{GAT} is the current state-of-the-art method. Further, we also include the results of the iterative classification algorithms~(ICA~\cite{lu2003link}) from~\cite{kipf2016semi}. 
\subsection{Results}

\subsubsection{Semi-Supervised Classification}
\label{sec:results}
\noindent
\textbf{Citation Networks. }The results for all the above-mentioned baseline methods and our two proposed models, GPGC-\emph{small} and GPGC-\emph{big}, are summarized in Table~\ref{tb:results}, where the best accuracy is highlighted in bold font. Since all the methods use the same data splits as indicated in~\cite{yang2016revisiting}, we cite the results of all the baseline methods from~\cite{ng2018bayesian}, which also reports results from~\cite{monti2017geometric} and~\cite{kipf2016semi}. We select the hyper-parameter with the best performance on the validation datasets, and then report the accuracy on the testing data in Table~\ref{tb:results}. From this table, we can observe that our GPGC-\emph{big} outperforms all other baseline methods on Cora and Pubmed, and is very competitive with GAT on Citeseer. %

We observe that GPGC-X and GGP-X significantly outperform other methods on all three datasets, which demonstrates the superiority of non-deep learning based methods on the task with few labeled samples.  
We also observe that GPGC-X outperforms GGP-X on all three datasets. This indicates that deep learning does have a stronger inference ability than statistical methods. 

\noindent
\textbf{Image Datasets. }From the results obtained for citation networks, we can observe that deep learning based methods perform better than non-deep learning ones. For ease of comparison on image datasets, we compare our methods with two deep learning models~(GCNs and GAT) and one GP-based model, GGP. Table~\ref{tb:image_results} summarizes the results of all these methods on MNIST, SVHN, and CIFAR10. We note that our proposed methods GPGC-\emph{small} and GPGC-\emph{big} obtain the best performance on all three datasets. We also observe that all methods achieve high accuracy on MNIST, with less variation between their results compared to those obtained on SVHN and CIFAR10. This is because, as can be seen from the t-SNE in Figure~\ref{fig:mnist}, all the classes in MNIST are distinctly clustered with little overlap. Thus, it is easy for algorithms to correctly classify them. 
When the data becomes more difficult to distinguish, the overall performance of all the methods deteriorates. 
Meanwhile, the improved classification accuracy of our methods over other models becomes more significant. 
This, in turn, indicates that our methods can better capture spatial correlations. 

\begin{table}[tb]
\caption{Results for Cora, Citeseer, and Pubmed}
\label{tb:results}
\begin{center}
\begin{tabular}{c|ccc}
\clineB{1-4}{2.5}
Method & Cora & Citeseer & Pubmed \\ \hline
LP~\cite{zhu2003semi} & 68.0\% & 45.3\% & 63.0\% \\
SemiEmb~\cite{weston2012deep} & 59.0\% & 59.6\% & 71.1\% \\
ManiReg~\cite{belkin2006manifold} & 59.5\% & 60.1\% & 70.7\% \\
DeepWalk~\cite{perozzi2014deepwalk} & 67.2\% & 43.2\% & 65.3\% \\
Planetoid~\cite{yang2016revisiting} & 75.7\% & 64.7\% & 77.2\% \\
ICA~\cite{zhu2003semi} & 75.1\% & 69.1\% & 73.9\% \\
DCNN~\cite{yang2016revisiting} & 76.8\% & - & 73.0\% \\
GCNs~\cite{kipf2016semi} & 81.5\% & 70.3\% & 79.0\% \\
MoNet~\cite{monti2017geometric} & 81.7\% & - & 78.8\% \\
GGP~\cite{ng2018bayesian} & 80.9\% & 69.7\% & 77.1\% \\ 
\textbf{GAT}~\cite{gat} & 83.0\% & \textbf{72.5}\% & 79.0\% \\ 
GPGC-\emph{small}~(Ours) & 81.2\% & 72.0\% & 78.0\% \\
GPGC-\emph{big}~(Ours) & \textbf{83.6\%} & 72.1\% & \textbf{79.6\%} \\ \hline
GGP-X~\cite{ng2018bayesian} & 84.7\% & 75.6\% & 82.4\% \\ 
GPGC-X~(Ours) & \textbf{86.0\%} & \textbf{76.9}\% & \textbf{83.8\%} \\
\clineB{1-4}{2.5}
\end{tabular}
\end{center}
\label{default}
\end{table}%

\begin{table}[tbp]
\caption{Results on MNIST, SVHN, and CIFAR10}
\label{tb:image_results}
\begin{center}
\begin{tabular}{c|ccc}
\clineB{1-4}{2.5}
Method & MNIST & SVHN & CIFAR10 \\ \hline
GCNs~\cite{kipf2016semi} & 98.6\% & 88.6\% & 70.2\% \\
GAT~\cite{gat} & 98.7\% & 88.4\% & 64.9\% \\ 
GGP~\cite{ng2018bayesian} & 98.6\% & 83.9\% & 70.2\% \\ 
GLCN~\cite{jiang2019semi} & 94.3\% & 79.1\% & 66.7\% \\ 
GPGC-\emph{small}~(Ours) & 98.8\% & \textbf{89.1\%} & 70.9\% \\
GPGC-\emph{big}~(Ours) & \textbf{98.9\%} & 88.8\% & \textbf{72.1\%} \\ \hline
GGP-X~\cite{ng2018bayesian} & 98.6\% & 88.5\% & 70.1\% \\
GPGC-X~(Ours) & \textbf{98.9\%} & \textbf{89.1\%} & \textbf{72.6\%} \\ \clineB{1-4}{2.5}
\end{tabular}
\end{center}
\label{default}
\end{table}%

\subsubsection{Running Time Efficiency}
\label{sect:runtime}

As described in Section~\ref{sec:eg_gcgp}, the computational complexity of our kernel matrix is the sum of its five steps. In Step 1, we first conduct a matrix multiplication between $\hat{\mA}$ and $\mX$, which incurs a computational cost of $\mathcal{O}(N|\epsilon|)$, where $|\epsilon|$ is the maximum degree of $\hat{\mA}$. Then, computing the covariance matrix $\mK(\hat{\mX}^{(0)}, \hat{\mX}^{(0)})$ costs $\mathcal{O}(N^2)$. 
Next, the time complexity for Step 2 is $\mathcal{O}(1)$, which is the same for Step 5. 
Following Equations~\ref{eq:acos_theta1} and~\ref{eq:acos_kernel1}, the total time complexity of Steps 2 and 3 is $\mathcal{O}(N^2)$. The most time consuming component comes from Step 4, which costs $\mathcal{O}(N^2)$ per element, as indicated by Equation~\ref{eq:gcn_f1}. However, this computation can be significantly reduced to $\mathcal{O}({|\epsilon|}^2)$ due to the sparsity of $\hat{\mA}$. Therefore, the total time complexity is $\mathcal{O}({N^2|\epsilon|}^2)$.

To demonstrate the time efficiency of our proposed method, we conduct experiments to compare the training and testing times of GCNs, GGP, GAT, and GPGC-\emph{big}. For fair comparison, all the methods are implemented in Python and run on a single CPU core without any GPUs. The results are reported in Table~\ref{Train_Test_time}. We can clearly observe that GPGC-\emph{big} has the lowest training time on graph datasets~(requiring around 1~s), e.g., Cora and Citeseer. Even on Pubmed, the training time of GPGC-\emph{big}~(46.2~s) is very close to that of GCNs~(43.8~s). Finally, it is also worth noting that GGP and GAT have much longer training and testing times compared to GCNs and GPGC-\emph{big}, which again further demonstrates the superiority of our proposed methods in terms of time efficiency. 

\begin{table}[t]
    \centering
    \small
    \caption{Training/Testing Time Comparison~(in seconds)}
    \label{Train_Test_time}
    \begin{tabular}{c|c|cccc}
    \clineB{1-6}{2.5}
        & Mode & GCNs & GGP & GAT & GPGC-\emph{big} \\
        \hline
       \multirow{2}{*}{Cora} 
       &\multicolumn{1}{c|}{Train} & 4.1 & 99.1 & 862.4 & \textbf{1.0} \\
        & \multicolumn{1}{c|}{Test} & 0.01 & 3.51 & 0.35 & 0.04 \\ \hline
        
        \multirow{2}{*}{Citeseer} &\multicolumn{1}{c|}{Train}  &5.9  & 1611.9  & 2073.0 & \textbf{1.4} \\
        & \multicolumn{1}{c|}{Test}  &0.01  & 7.12 & 1.02 & 0.04 \\ \hline
        
        \multirow{2}{*}{Pubmed} &\multicolumn{1}{c|}{Train} & \textbf{43.8} & 2429.0  & 1721.1 & 46.2 \\
        & \multicolumn{1}{c|}{Test} & 0.10 & 111.06 & 1.22 & 0.09 \\
        \clineB{1-6}{2.5}
    \end{tabular}
\end{table}

\begin{table}[t]
\caption{Results for Different Units of GCNs}
    \centering
        \begin{tabular}{c|cccc}%
        \clineB{1-5}{2.5}
         $Num_{Hidden}$ & 16 & 64 & 256 & +$\infty$ \\ \hline
Cora  	 & 81.5\%   & 82.2\%  & 83.0\% & \textbf{83.6\%} \\ 
Citeseer & 69.7\%   & 70.9\%  & 71.0\% & \textbf{72.1\%} \\
Pubmed   & 79.0\%   & 79.3\%  & 79.3\% & \textbf{79.6\%} \\
        \clineB{1-5}{2.5}
        \end{tabular}
    \label{tabl:diff_num_layers}
\end{table}

\subsubsection{Effect of GCN Units Number}

To demonstrate the superiority of infinitely wide GCNs, we conduct an experiment by increasing the number of hidden units. Table~\ref{tabl:diff_num_layers} reports the classification accuracy of GCNs with 16, 64, and 256 hidden units, on three citation datasets. 
For better comparison, we show the results of GPGC-\emph{big} in column ``$+\infty$" in~Table~\ref{tabl:diff_num_layers}. 
We consistently observe, across all three datasets, that the classification accuracy improves with an increasing number of hidden units. 
We also observe that the best classifications results are always obtained at $+\infty$, which is consistent with the underlying trend discussed above. 

\subsubsection{Effect of Different Amounts of Labeled Samples}
\begin{table}[h!t]
\caption{Results of Different Amounts of Labeled Samples}
    \centering
        \begin{tabular}{c|c|ccc}%
        \clineB{1-5}{2.5}
         & Model & 200 & 600  & 1000 \\ \hline
\multirow{3}{*}{MNIST} & GCNs 	
& 98.4\%    & 98.6\%  & 98.6\%  \\ 
& GPGC-\emph{big}	
& \textbf{98.5\%}   & \textbf{98.8\%}  & \textbf{98.9\%} \\ \clineB{2-5}{1.0}
& GPGC-X	
& \textbf{98.6\%}   & \textbf{98.8\%}  & \textbf{99.0\%} \\ \hline
\multirow{3}{*}{SVHN} & GCNs 	
& 87.7\%    & 88.3\%  & 88.6\%  \\ 
& GPGC-\emph{big}	
& \textbf{88.1\%}   & \textbf{88.7\%}  & \textbf{88.8\%} \\ \clineB{2-5}{1.0}
& GPGC-X	
& \textbf{88.6\%}   & \textbf{89.0\%}  & \textbf{89.1\%} \\ 
\hline
\multirow{2}{*}{CIFAR10} & GCNs 	
& 67.4\%    & 69.4\%  & 70.2\%  \\ 
& GPGC-\emph{big}	
& \textbf{67.3\%}    & \textbf{70.4\%}  & \textbf{72.1\%} \\ \clineB{2-5}{1.0}
& GPGC-X	
& \textbf{72.1\%}   & \textbf{72.1\%}  & \textbf{72.6\%} \\ 
        \clineB{1-5}{2.5}
        \end{tabular}
    \label{tabl:diff_num_labels}
\end{table}

In this experiment, we study the problem of semi-supervised classification with different amounts of labeled data. We consider identical experimental settings for GCNs, GPGC-\emph{big} and GPGP-X on the image datasets, MNIST, SVHN, and CIFAR10. We vary the number of labeled images as 200, 600, and 1000 for each dataset. As shown in Table~\ref{tabl:diff_num_labels}, the performance of GCNs, GPGC-\emph{big} and GPGP-X increases when more labeled data is available. This follows the intuition that, with more labeled data, the decision boundaries for different classes align better. Thus, the classification results are improved. When the number of labeled images goes from 600 to 1000, we note that the performance on MNIST and SVHN seems to stop increasing, while the improvements on CIFAR10 are still significant, i.e., GCNs boost the accuracy from 58.1\% to 58.6\%, and GPGC-\emph{big} improves it from 58.9\% to 60.0\%. 
This suggests that the labeled data plays an important role when the data cannot be clustered distinctly, for example, the t-SNE for CIFAR10 in Figure~\ref{fig:cifar10}. 
Further, we observe that GPGC-\emph{big} consistently achieves better performance than GCNs with the same number of labeled images, on all the datasets. This in turn indicates that GPGC-\emph{big}, which is two layers GCNs with an infinite number of hidden units, is always better than the one with a limited number.

\section{Conclusions}
In this paper, we have shown that a GCN with infinitely wide layers is equivalent to a GP. We also derived an analytical form of computing the covariance matrix when the activation function is restricted being~\emph{ReLU}. We then re-framed the graph-based semi-supervised classification into a multi-output GP regression problem, deriving the corresponding mean and covariance matrix for the GP posterior. 
Extensive experiments demonstrated that the proposed GPGC-\emph{small} and GPGC-\emph{big} outperform all the other baselines by a clear margin. 
By taking advantage of the sparsity of the adjacent matrix and the analytical form derivation, we noticed that our proposed method is easy to train, taking only a few seconds. 
We also demonstrated that, with an increasing number of units in GCNs, the performance improves accordingly, which in turn shows the superiority of infinitely wide GCNs over finite ones.

{\small
\bibliographystyle{plain}
\bibliography{1st}
}

\end{document}